\documentclass{article}

\usepackage{arxiv}

\usepackage[utf8]{inputenc} 
\usepackage[T1]{fontenc}    
\usepackage{hyperref}       
\usepackage{url}            
\usepackage{booktabs}       
\usepackage{amsfonts}       
\usepackage{times}  
\usepackage{helvet}  
\usepackage{courier}  
\usepackage{graphicx} 
\usepackage{booktabs} 
\usepackage{array}    
\usepackage{amsmath}  
\usepackage{amssymb}  
\usepackage{amsmath}        
\usepackage{nicefrac}       
\usepackage{microtype}      
\usepackage{lipsum}
\usepackage{algorithm}
\usepackage{algorithmic}
\graphicspath{ {./images/} }

\title{ECG-aBcDe: Overcoming Model Dependence, Encoding ECG into a Universal Language for Any LLM}

\author{
 Yong Xia
 \And
 Jingxuan Li 
 \And
  YeTeng Sun
 \And
  Jiarui Bu 
 \And
}

\begin{document}
\maketitle
\begin{abstract}
Large Language Models (LLMs) hold significant promise for electrocardiogram (ECG) analysis, yet challenges remain regarding transferability, time-scale information learning, and interpretability. Current methods suffer from model-specific ECG encoders, hindering transfer across LLMs. Furthermore, LLMs struggle to capture crucial time-scale information inherent in ECGs due to Transformer limitations. And their black-box nature limits clinical adoption. To address these limitations, we introduce ECG-aBcDe, a novel ECG encoding method that transforms ECG signals into a universal ECG language readily interpretable by any LLM. By constructing a hybrid dataset of ECG language and natural language, ECG-aBcDe enables direct fine-tuning of pre-trained LLMs without architectural modifications, achieving "construct once, use anywhere" capability. Moreover, the bidirectional convertibility between ECG and ECG language of ECG-aBcDe allows for extracting attention heatmaps from ECG signals, significantly enhancing interpretability. Finally, ECG-aBcDe explicitly represents time-scale information, mitigating Transformer limitations. This work presents a new paradigm for integrating ECG analysis with LLMs. Compared with existing methods, our method achieves competitive performance on ROUGE-L and METEOR. Notably, it delivers significant improvements in the BLEU-4, with improvements of 2.8 times and 3.9 times in in-dataset and cross-dataset evaluations, respectively, reaching scores of 42.58 and 30.76. These results provide strong evidence for the feasibility of the new paradigm.
\end{abstract}


\section{Introduction}

\begin{figure*}[t]
\centering
\includegraphics[width=0.95\textwidth]{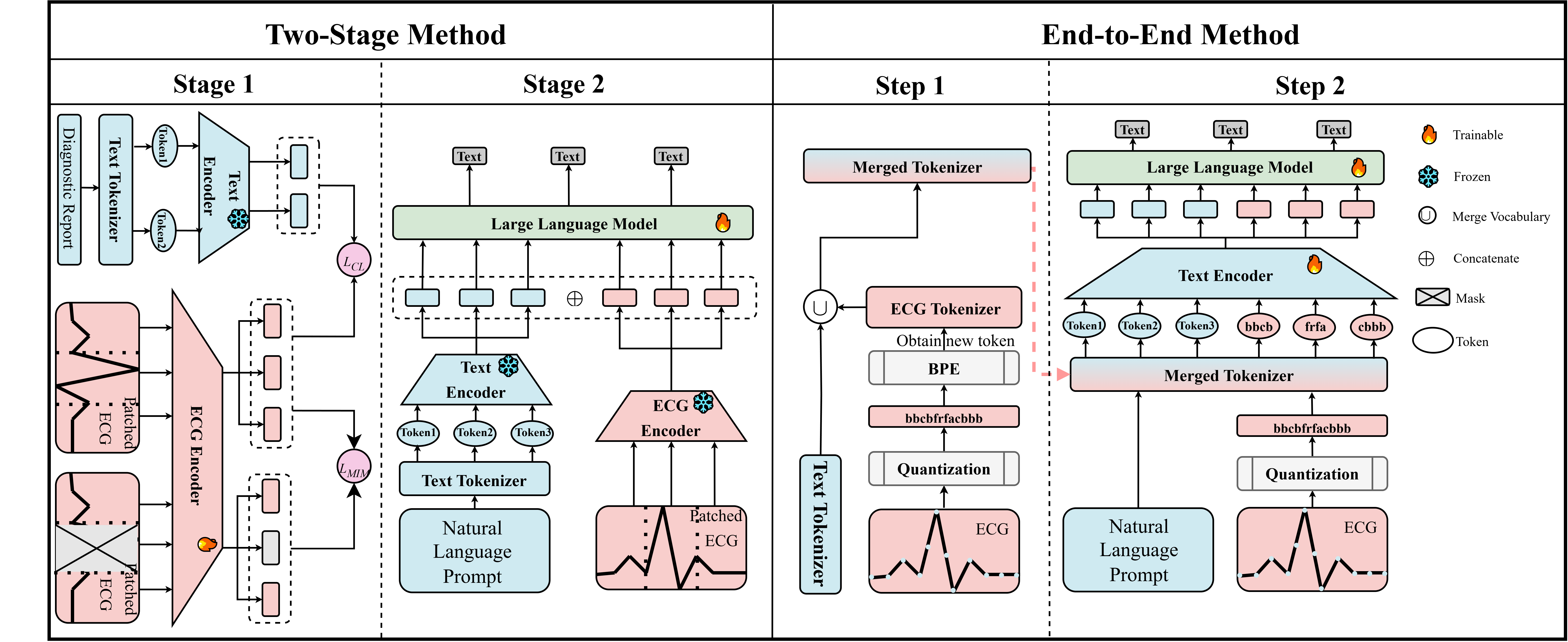} 
\caption{Comparison between Two-Stage Methods and End-to-End Methods. In scenarios where methods trained on earlier LLMs want to be applied to more advanced LLMs, both two-stage and end-to-end methods require model-specific adaptations before fine-tuning. Specifically, two-stage methods involve retraining the ECG encoder to align with the text encoder of the new LLM, whereas end-to-end methods require integrating the ECG tokenizer with the tokenizer of the target LLM.}
\label{difference}
\end{figure*}

Electrocardiograms (ECGs) are widely used in clinical practice due to their non-invasive and rapid nature. They significantly reduce diagnosis and treatment time for acute conditions such as myocardial infarction and ventricular fibrillation \cite{DLECGLiterature}, while also aiding early detection of asymptomatic high-risk individuals during routine check-ups \cite{ECG_Mortality}. According to the World Health Organization, nearly 200 million ECGs are performed globally each year \cite{WHO2019}, highlighting the importance of related research.

With the rapid advancement of deep learning in time-series analysis, numerous models have been introduced into the ECG analysis, including convolutional neural networks \cite{CNNSVM,2dCNN}, long short-term memory networks \cite{LSTMA}, and Transformers \cite{CWTransformer}. State-of-the-art methods have achieved classification accuracies as high as 99.38\% \cite{CWTransformer}. Despite these impressive results in ECG classification tasks, current methods typically produce a single categorical label as output, which limits their applicability in real-world automated diagnosis scenarios that demand richer information and interpretability.

Large language models (LLMs) produce natural language outputs, which align closely with human reading habits and allow classification tasks to extend to more complex queries, such as symptom description. These capabilities make LLMs well suited for automated clinical diagnosis. As a result, integrating ECG with LLMs has become a research focus. As shown in Figure \ref{difference}, current methods for ECG and LLM integration can be divided into two categories: \textbf{two-stage methods} and \textbf{end-to-end methods}. Early studies adopted the two-stage method, where ECGs are first segmented into patches and encoded using an ECG encoder aligned with the LLM text encoder via self-supervised \cite{selfa,selfb} and contrastive learning \cite{MERL}. During ECG encoder training, masked image modeling loss ($L_{MIM}$) and contrastive loss ($L_{CL}$) are commonly used for joint optimization. Despite their strong performance, two-stage methods suffer from limited training efficiency and poor cross-model transferability, mainly due to reliance on large ECG datasets and specific LLM text encoders. To overcome these limitations, recent work has proposed an end-to-end method \cite{ECG_Byte}. Its core idea is to quantize ECG signals into lowercase character sequences, further compressed via Byte Pair Encoding (BPE) \cite{BPE}. The resulting ECG token sequence is directly concatenated with natural language instructions and jointly input into the LLM, eliminating the need for a separate ECG encoder and enabling unified end-to-end training for both representation learning and downstream reasoning.

Although end-to-end methods address some limitations of previous two-stage methods, challenges remain in combining ECG with LLMs. First, while end-to-end methods address limitations of two-stage methods, they still require model-specific adaptation before fine-tuning, as shown in Figure \ref{difference}. Second, both time-scale and morphological features are equally important in ECG analysis. However, LLMs have significantly weaker ability to learn time-scale features compared to morphological ones. Existing methods have paid limited attention to this issue. For example, normal RR intervals range from 0.6 to 1.0s, while patients with atrial fibrillation may have RR intervals exceeding 2.5s. Some studies improve performance by concatenating important time-scale features like RR intervals with ECG embeddings \cite{RR1}, but they do not explore why deep learning struggles to learn such features directly. It was noted that Google identified inherent limitations of the Transformer architecture in addressing counting tasks \cite{whencountn}.  This inspired us to relate counting the number of lowercase ‘a’s between two ‘R’s in “RaaabaaR” to learning temporal features like RR intervals. The Transformer’s weakness in counting manifests in ECG analysis as LLMs failing to learn time-scale information, such as QRS duration and RR intervals—even when trained on massive data and with large parameter sizes. This limitation results in poor analysis of diseases that rely heavily on time-scale features, such as atrial fibrillation.

To address the limitations of existing methods, this work departs from the conventional focus on improving model architectures. Instead, it draws inspiration from clinical diagnostic practices to develop an efficient encoding and representation method for ECGs. In clinical diagnosis, ECG analysis relies not on fixed-length patches but on the signal’s intrinsic physiological structure, which is segmented into key components such as the P wave, QRS complex, and T wave. Diagnosis is performed by analyzing their waveform characteristics and time-scale information. For example, prolongation of the QRS complex is a typical indicator of bundle branch block, while irregular variations in RR intervals may signify atrial fibrillation. Motivated by this, a novel ECG encoding method named ECG-aBcDe is proposed, which aligns with clinical diagnostic logic. This method extracts key points and their corresponding inter-point intervals from ECG, and through quantization, converts the continuous ECG into a discrete symbolic sequence alternating between key points and their corresponding inter-point intervals. This representation captures both waveform characteristics and explicitly encodes temporal information, thereby better aligning with the clinical diagnostic reasoning process. Compared to existing methods, it enables a more effective and concise semantic representation of ECGs, enhancing both model interpretability and performance.
In summary, the contributions of this study are as follows:

\begin{itemize}
    \item This study proposes an LLM-agnostic ECG encoding method, ECG-aBcDe, which converts ECG into an ECG language resembling natural language. Any LLM can comprehend the ECG language through fine-tuning on the constructed dataset comprising both ECG and natural language, overcoming the limitation of existing methods that require adaptation to specific model, simplifying the experimental workflow.
    \item This study identifies the limitations of LLMs in representing time-scale information in ECG signals and proposes a simple and effective solution.
    \item This study provides verifiable evidence for the outputs generated by LLMs through attention weight visualization on reconstructed ECG signals decoded by ECG language, thereby effectively enhancing the interpretability of the results.
    \item The proposed method achieved BLEU-4 scores that were 2.8 times and 3.9 times higher than the best baseline on within-dataset and cross-dataset evaluations, validating our advantages in generation quality and generalization capability.
\end{itemize}

\section{Related Work}
\subsection{Key Point Detection in Time Series  }

Key point detection methods for time series mainly fall into three categories: statistical methods, objective function optimization methods, and machine learning methods. Statistical methods \cite{CUSUM,ZScore} typically rely on thresholds, sliding windows, and cumulative statistics to identify key points. Objective function optimization methodes define explicit objective functions and solve for key points automatically \cite{opta,optb}. Machine learning methods generally achieve better accuracy but are less efficient. For instance, the common k-means clustering has a time complexity of $O(N\cdot K)$ \cite{kmeans}. Considering both performance and efficiency in ECG key point detection, this work adopts the L1 trend filtering \cite{L1Trend}, which solves for key points efficiently. The optimization objective is defined as:
\begin{equation}
\min_{x \in \mathbb{R}^n}  \frac{1}{2} \sum_{i=1}^{n} \left( y_i - x_i \right)^2 + \lambda \sum_{i=2}^{n-1} \left| x_{i-1} - 2x_i + x_{i+1} \right|
\label{eq1} 
\end{equation}
,where $y$ is the original signal, $x$ is the optimization variable, and  $\lambda$ is the regularization parameter for the L1 norm of the second-order difference term. This objective captures piecewise smooth trends while preserving discontinuities at sudden signal changes, enabling effective key point detection. Moreover, the problem is convex and guarantees a global optimum. It can be efficiently solved in near-linear time by ECOS \cite{ECOS}, balancing accuracy and efficiency.

\subsection{Counting Problems in Large Language Models}
  
In the point-wise encoded ECG representation \cite{ECG_Byte}, the ECG is converted into a character sequence such as “raaaar”, where each character corresponds to the voltage value at a sampling point. In this representation, key physiological features like the RR interval are formulated as ”How many characters occur between two 'r's?" Thus, enabling LLMs to capture time-scale information essentially requires them to perform counting.

Unfortunately, despite their impressive performance in natural language processing, LLMs remain limited in counting tasks. For instance, the “strawberry test” reveals that models often fail to correctly count the number of 'r's in the word “strawberry”. Several studies have explored the underlying causes of this limitation. Reference \cite{strawberry} attributes part of the issue to the tokenizer: LLMs attend at the token level (e.g., “staw” and “berry”) and struggle to capture character-level details. Beyond tokenization, It has been suggested that attention normalization may dilute counting information \cite{tfla}. Even if the model correctly attends to all three instances of 'r', normalization can average out the attention weights, leading to a loss of count-specific signals. To address this issue, some studies propose removing attention normalization to improve counting performance \cite{tfla}. It has also been shown that Transformers can theoretically perform counting when the vocabulary size is smaller than the embedding dimension, with corresponding implementations provided \cite{whencountn}. More recently, chain-of-thought (CoT) prompting has been demonstrated to significantly improve counting accuracy, though its high computational cost limits practical use \cite{ACL2025}. In summary, prior research reveals the limitations of LLMs in counting tasks and offers several solutions, yet challenges in efficiency and practicality remain, warranting further investigation.

\section{Method}
\begin{figure*}[ht]
\centering
\includegraphics[width=0.85\textwidth]{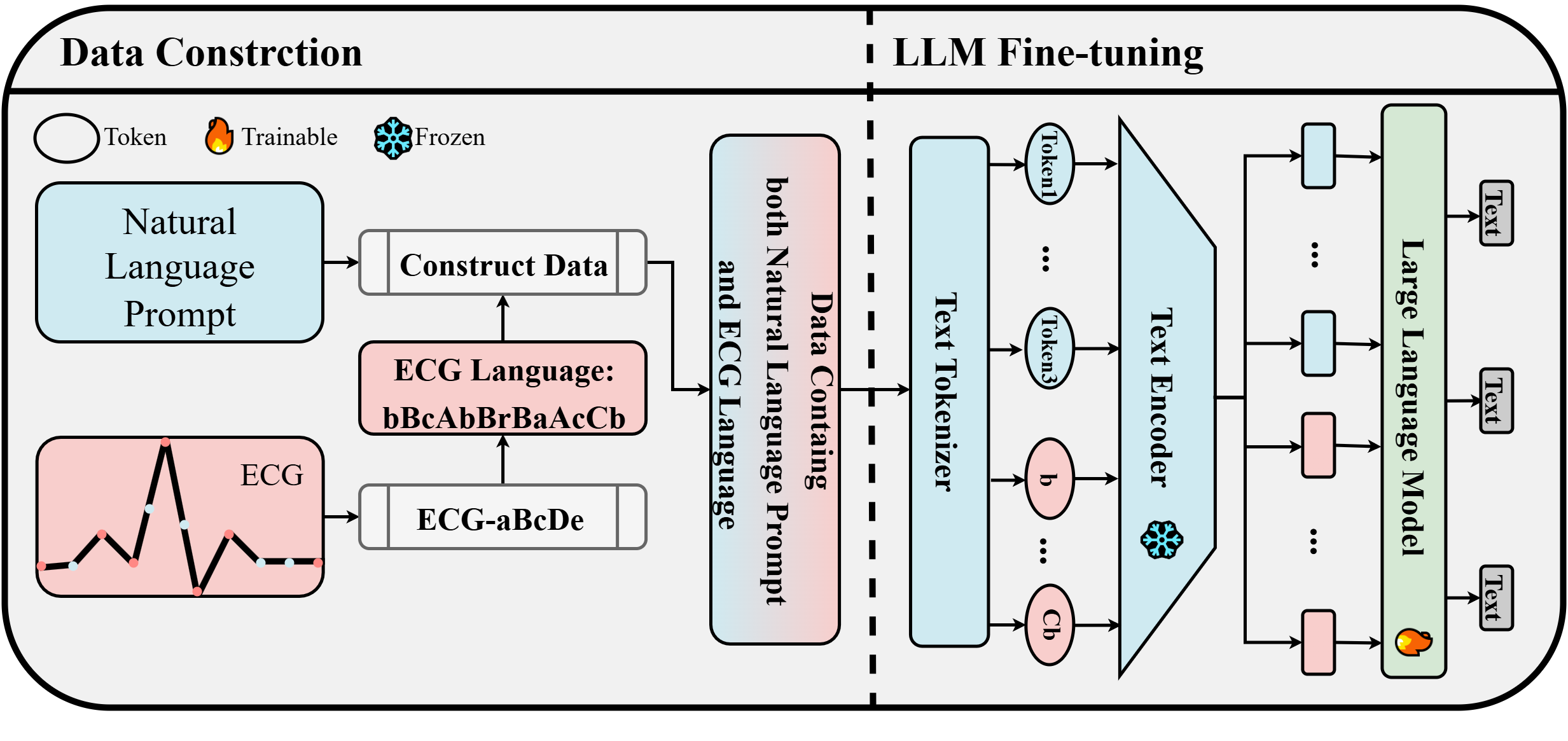} 
\caption{Overview of our method. Compared to the method in Figure \ref{difference}, our method fully decouples the task of ECG analysis into two independent parts: the construction of data containing both ECG language and natural language, and the subsequent fine-tuning of the LLM.}
\label{overview}
\end{figure*}

As shown in Figure \ref{overview}, the proposed method achieves comprehension of ECG by LLMs through two independent stages: \textbf{Data Construction} and \textbf{LLM Fine-Tuning}. Specifically, the data construction stage transforms continuous ECG into a discrete symbolic sequence termed ECG language, and constructs training samples by pairing ECG language with natural language prompts. During fine-tuning, the LLM is trained using sequences formed by concatenating ECG language with natural language prompts. To retain linguistic capabilities while adapting to the new modality, only the upper layers of the LLM are updated, with the text encoder frozen. This enables efficient integration of ECG semantics without compromising general language understanding. Ultimately, the model generates accurate textual analysis results.

\subsection{ECG-aBcDe}
This section systematically describes the implementation details of ECG-aBcDe, which enables bidirectional mapping between ECG signals and ECG language.

\subsubsection{Preprocessing}
During preprocessing, 12-lead ECG signals are reordered to [I, II, III, aVL, aVR, aVF, V1–V6] for lead consistency. Denoising is applied using a bidirectional notch filter at 50/60Hz (Q=30), followed by a fourth-order Butterworth bandpass filter (0.5–100Hz) and a bidirectional fourth-order high-pass filter (cutoff: 0.05Hz) to remove baseline wander. Daubechies-6 wavelet denoising (level 4) is then performed with soft thresholding based on the median absolute deviation of detail coefficients. Finally, signals are resampled to a frequency $F$ = 250Hz, yielding $X \in \mathbb{R}^{N \times 12 \cdot (F \cdot T)}$, where $N$ is the sample count and $T$ denotes duration in seconds.

\subsubsection{Encoding and Decoding}

Prior to encoding, general statistical information is extracted from the training set. Specifically, all ECG voltage values $x$ across sampling points are sorted to form the dataset $\mathcal{D}_V$, from which the 1st and 99th percentiles is denoted as $x_1$ and $x_{99}$, respectively. Let $V_L = x_1$ and $V_H = x_{99}$, which serve as the lower and upper bounds of the ECG voltage. The closed interval $[V_L, V_H]$ is uniformly divided into 24 equal-length subintervals by 23 segmentation points $V_1,\ldots,V_{23}$. Define $V_0 = V_L$ and $V_{24} = V_H$, and form the ordered set \(V = [V_0, V_1, \ldots, V_{24}]\), which divides the voltage value range into 26 non-overlapping intervals.

Next, key points of the ECG signal are extracted using $L_1$ trend filtering \cite{L1Trend}, and the time intervals between adjacent key points are computed. All such intervals from the training samples are sorted to form a new dataset $\mathcal{D}_T$. This dataset is then divided equally into 26 subintervals based on the element count. The corresponding  segmentation points form the ordered set \(T = [T_0, T_1, \ldots, T_{24}]\).
Note that the quantization strategies differ for voltage values and time intervals: voltage values are partitioned by magnitude (equal-width intervals), while time intervals are partitioned by element frequency (equal-count intervals). This method ensures a more uniform distribution after quantization. For convenience in subsequent processing, the sets are extended by appending an upper bound to each, resulting in \(V = [V_0, V_1, \ldots, V_{24}, +\infty]\) and \(T = [T_0, T_1, \ldots, T_{24}, +\infty]\).

With the sets $V$ and $T$ obtained, the quantization process can be formalized via mapping functions. Let the lowercase alphabet be defined as $\mathcal{A}_L = \{a, b, \ldots, z\}$. A surjective mapping $f_V: \mathbb{R} \rightarrow \mathcal{A}_L$ is defined as:
\begin{equation}
f_V(x) = \mathcal{A}_L[i], \quad \text{where } i = \max\{j \in [0, 25] \mid x \leq V_j\}.
\label{eq2} 
\end{equation}
That is, each sampled point x of the ECG signal is mapped to a symbolic letter representing its assigned quantization level. The inverse mapping $g_V: \mathcal{A}_L \rightarrow \mathbb{R}$ satisfies:
\begin{equation}
g_V(a_L) = V[i], \quad \text{where } i \text{ is the index of } a_L \text{ in } \mathcal{A}_L.
\label{eq3} 
\end{equation}

This maps each lowercase letter $a_L$ to the upper bound of the interval it represents. For the case of $a_L = z$, we use $V_{25} = 2V_{24}$ in implementation to avoid infinite values. Similarly, for time intervals, an uppercase alphabet $\mathcal{A}_U = \{A, B, \ldots, Z\}$ is defined, and analogous mapping functions $f_T$ and $g_T$ are constructed.

Using $f_V$, $g_V$, $f_T$, and $g_T$, bidirectional conversion between real-valued ECG signals and symbolic ``ECG language'' is achieved.

Based on the above quantization, the encoding procedure of ECG-aBcDe is as follows. First, $L_1$ trend filtering extracts key points from the ECG signal. The original signal is transformed into an alternating sequence of key point values and inter-point intervals. Each key point value is quantized into a lowercase letter using $f_V$, and each interval is quantized into an uppercase letter using $f_T$. The resulting character sequence (e.g., "aBcDe") forms the "ECG language." The pseudocode for encoding is given in Algorithm \ref{alg:encode}. This method enables any ECG signals to be transformed into the ECG language, serving as the basis for dataset construction.

\begin{algorithm}[htb]
\caption{ECG-aBcDe Encoding Pseudo Code}
\label{alg:encode}
\textbf{Input}: Single-lead ECG $x \in \mathbb{R}^{F\cdot T}$\\
\textbf{Output}: ECG language $x_{\text{lang}}$
\begin{algorithmic}[1]
\STATE $x_{\text{lang}} \leftarrow [\ ]$ 
\STATE $id \leftarrow \text{L1\_Trend\_Filter}(x)$ //The list of keypoint’s indices
\STATE $x_{key} \leftarrow x[id]$ //The list of keypoint
\FOR{$i \leftarrow 0$ to $|id| - 1$}
    \STATE $x_{\text{lang}}$.append($f_{V}(x_{key}[i])$)
    \STATE $x_{\text{lang}}$.append($f_{T}(id[i+1]-id[i])$)
\ENDFOR
\STATE \textbf{return} $x_{\text{lang}}$
\end{algorithmic}
\end{algorithm}

With the encoding mechanism established, the decoding process reconstructs the original ECG signal from the ”ECG language" using the previously defined mappings \(g_V\) and \(g_T\). Specifically, during decoding, these two mappings are applied to reconstruct both the key point magnitudes and their corresponding inter-point intervals in the ECG signal. The values of all intermediate sampling points between adjacent key points are then filled in using linear interpolation, thereby reconstructing the complete ECG waveform. The pseudocode for this decoding process is provided in Algorithm~\ref{alg:decode}. This decoding procedure enables the reconstruction of any ``ECG language'' back to ECG signal, providing the foundation for the interpretability of generation results.

\begin{algorithm}[htb]
\caption{ECG-aBcDe Decoding Pseudo Code}
\label{alg:decode}
\textbf{Input}: ECG language $x_{\text{lang}}$\\
\textbf{Output}: ECG $x \in \mathbb{R}^{F \cdot T}$
\begin{algorithmic}[1]
\STATE $x \leftarrow [\ ],\ m, n \leftarrow 0,\ x[0] \leftarrow g_{V}(x_{\text{lang}}[0])$
\FOR{$i \leftarrow 1$ to $\left\lfloor \frac{|x_{\text{lang}}|}{2} \right\rfloor$}
    \STATE $n \leftarrow m + g_{T}(x_{\text{lang}}[2i - 1])$
    \STATE $x[n] \leftarrow g_{V}(x_{\text{lang}}[2i])$
     \\//Linear interpolation
    \FOR{$j \leftarrow m + 1$ to $n - 1$} 
        \STATE $x[j] \leftarrow x[m] + \frac{j - m}{n - m} \cdot (x[n] - x[m])$
    \ENDFOR
    \STATE $m \leftarrow n$
\ENDFOR
\STATE \textbf{return} $x$
\end{algorithmic}
\end{algorithm}

\subsection{Data Construction}\label{data construction}

\begin{figure}[htbp]
    \centering
    \includegraphics[width=0.9\columnwidth]{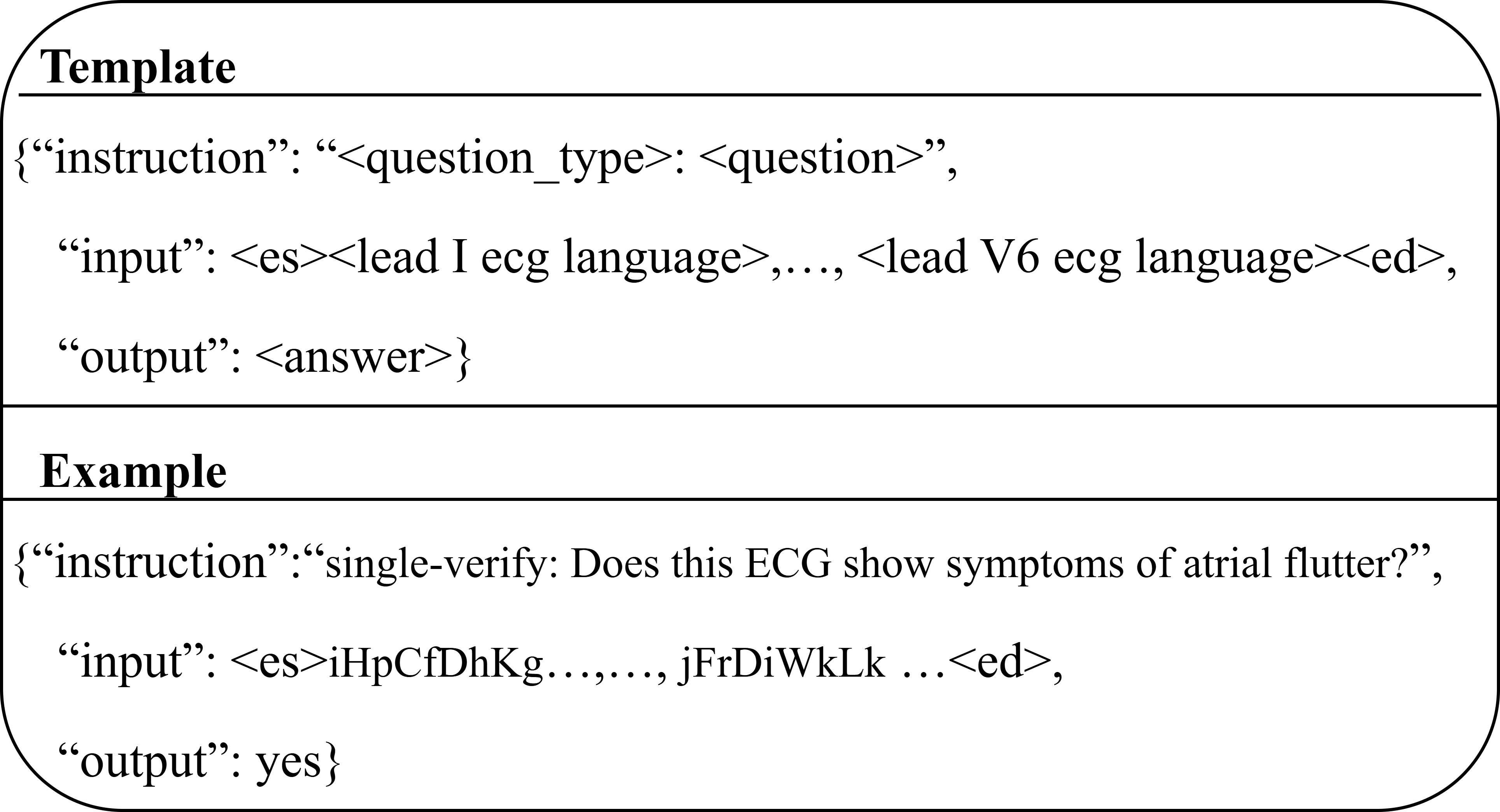} 
    \caption{Template and example of the data.}
    \label{fig:template}
\end{figure}

In this section, an ECG fine-tuning dataset compatible with any LLM is constructed based on the ECG language encoded by ECG-aBcDe and matched clinical question–answer pairs. The dataset template and an example are illustrated in Figure \ref{fig:template}. Specifically, the fields \textless{}question\_type\textgreater{}, \textless{}question\textgreater{}, and \textless{}answer\textgreater{} are automatically generated via the ChatGPT API ~\cite{ECGQA}, while the \textless{}es\textgreater{} and \textless{}ed\textgreater{} tokens in the “input” field denote the start and end of an ECG segment, respectively. The content between each pair of  \textless{}es\textgreater{} and \textless{}ed\textgreater{} contains the ECG language derived from the original 12-lead ECG signals through ECG-aBcDe encoding.

\subsection{Learning Objective}
For the dataset comprising ECG language, natural language instruction and output, the objective of supervised fine-tuning can be formalized as a masked autoregressive cross-entropy loss. Under this framework, each data sample consists of an instruction \( I^{(i)} \), an ECG language sequence \( x_{\text{lang}}^{(i)} \), and a target output \( Y^{(i)} \). A unified prompt sequence is constructed by combining the instruction and ECG language using a template, where the ECG sequence is wrapped by special tokens \textless{}es\textgreater and \textless{}ed\textgreater, yielding \( P^{(i)} = I^{(i)} \parallel\)   \textless{}es\textgreater \(x_{\text{lang}}^{(i)}\) \textless{}ed\textgreater, where the operator \(\parallel\) represents the concatenation of symbolic representations. This prompt is then concatenated with the target output to form the full input sequence to the model: \(S^{(i)} = P^{(i)} \parallel Y^{(i)} \). The training objective is to minimize the following loss function \( L_{\text{SFT}} \):

\begin{equation}
 L_{\text{SFT}}(\theta) = -\frac{1}{N} \sum_{i=1}^{N} \sum_{t = |P^{(i)}| + 1}^{|S^{(i)}|} \log P\left(s_t^{(i)} \mid s_{<t}^{(i)}; \theta\right),
\label{eq3:loss} 
\end{equation}
where the key element lies in the inner summation starting from \( t = |P^{(i)}| + 1 \), which implements a loss-masking mechanism. This ensures that the loss is computed only over the target output tokens \( Y^{(i)} \), while the prompt (including the instruction and ECG data) provides context to the model without contributing to gradient updates.


\section{Experiment}
\subsection{Setup}
\subsubsection{Baselines}
This study compares the proposed ECG-aBcDe with representative two-stage and end-to-end baseline methods. The baselines include D-BETA \cite{selfa}, which is based on contrastive learning; ECGBERT \cite{ECGBERT}, which adopts masked image modeling; MERL \cite{MERL}, a multimodal method integrating self-supervised learning; and the state-of-the-art end-to-end method ECG-Byte \cite{ECG_Byte}.
\subsubsection{Datasets and Metrics}
This study adopts two benchmark datasets, ECG-QA PTB-XL and ECG-QA MIMIC-IV \cite{ECGQA}, encompassing seven distinct types of questions: single choose, single query, single verify, comparison consecutive verify, comparison consecutive query, comparison irrelevant verify, and comparison irrelevant query. These categories are designed to comprehensively evaluate the model’s capabilities in ECG interpretation and question answering. To objectively assess the quality of generated responses, three widely used metrics—BLEU-4\cite{BLEU}, ROUGE-L\cite{Rouge} and METEOR\cite{Meteor} are employed to measure the semantic consistency and textual similarity between the model outputs and the reference answers.

\subsubsection{Implementation Details}
To ensure a fair comparison, all experiments were conducted using Llama 3.2-1B-Instruct \cite{Llama}. During training, the LoRA method \cite{lora} was employed with hyperparameters set to $r = 16$ and $\alpha = 32$. The model was fine-tuned using the AdamW optimizer with a learning rate of $1 \times 10^{-4}$ \cite{AdamW}, a cosine learning rate scheduler \cite{SGDR}, and 500 warm-up steps. The input sequence length was truncated to 4096, with a batch size of 2 and a total of one training epoch. During inference, Top-P sampling was set to 0.7, the temperature was 0.9, and the input sequence length was likewise truncated to 4096.

\subsection{Results}
To comprehensively evaluate the performance of ECG-aBcDe, this study conducted experiments under two scenarios: in-distribution evaluation on the source dataset and zero-shot cross-dataset transfer. The results are summarized in Table~\ref{tab:within} and \ref{tab:cross}, where bold indicates the best performance and underlines denote the second-best.
\begin{table}[ht]
\caption{Results of Within-Dataset}
\label{tab:within}
\centering
\scalebox{1.0}{ 
\begin{tabular}{@{}lcclcc@{}}
\toprule
\multicolumn{1}{l}{\textbf{Method}} & 
\multicolumn{1}{c}{\textbf{Tr. Data}} & 
\multicolumn{1}{c}{\textbf{Inf. Data}} & 
\multicolumn{1}{c}{\textbf{BLEU-4}} & 
\multicolumn{1}{c}{\textbf{Rouge-L}} & 
\multicolumn{1}{c}{\textbf{Meteor}} \\
\midrule
D-BETA    &  &  & \underline{15.14} & 46.71 & \textbf{29.64} \\
ECGBERT   &  &  & 12.04 & 40.44 & 24.84 \\
MERL   & \multicolumn{1}{c}{PTB-XL} & PTB-XL & 11.27 & 38.83 & 24.03 \\
ECG-Byte    &  &  & 13.93 & \underline{47.08} & 29.17 \\
ECG-aBcDe   &  &  & \textbf{42.88} & \textbf{50.55} & \underline{29.32} \\
\bottomrule
\end{tabular}
}
\end{table}

As shown in Table \ref{tab:within}, in the within-dataset evaluation, ECG-aBcDe achieves comparable performance to state-of-the-art methods in terms of ROUGE-L and METEOR, while significantly outperforming them in BLEU-4 with a score of 42.58. These results demonstrate that ECG-aBcDe outperforms existing methods in capturing both lexical and phrasal correspondences, consequently producing higher-quality text with closer alignment to reference descriptions.

\begin{table}[ht]
\caption{Results of Zero-Shot Cross-Dataset Transfer}
\label{tab:cross}
\centering
\scalebox{1.0}{
\begin{tabular}{@{}lccccc@{}}
\toprule
\textbf{Method} & \textbf{Tr. Data} & \textbf{Inf. Data} & \textbf{BLEU-4} & \textbf{Rouge-L} & \textbf{Meteor} \\
\midrule
D-BETA &  &  & 5.10 & 22.77 & 14.63 \\
ECGBERT &  &  & 7.68 & \underline{35.77} & \underline{22.32} \\
MERL & PTB-XL & MIMIC-IV & 7.39 & 28.33 & 18.59 \\
ECG-Byte &  &  & \underline{7.86} & 35.01 & 21.49 \\
ECG-aBcDe &  &  & \textbf{30.76} & \textbf{38.57} & \textbf{22.56} \\
\bottomrule
\end{tabular}
}
\end{table}

And in the more challenging zero-shot cross-dataset transfer setting, Table \ref{tab:cross} shows that ECG-aBcDe exhibits strong generalization ability. Although the overall performance of all models declines in this setting, ECG-aBcDe consistently outperforms others across all evaluation metrics, validating its robustness and adaptability to out-of-distribution data. In summary, the experimental results clearly demonstrate that, compared to prior methods, ECG-aBcDe is more effective in capturing critical features of electrocardiographic signals and maintains strong generalization capabilities when dealing with previously unseen data distributions.

To further validate the generalizability of ECG-aBcDe, we fine-tuned several large language models based on the Llama-Factory framework \cite{llamafactory}. Specifically, we conducted experiments on Gemma3-1B-Instruct \cite{gemma3}, Llama 3.2-1B-Instruct \cite{Llama}, and Qwen2.5-0.5B-Instruct \cite{qwen2.5}. The results presented in Table \ref{tab:vllms} indicate that, even across different large language models, all models are capable of effectively understanding and processing the ECG language when trained with our constructed instruction tuning dataset.

\begin{table}[ht]
\caption{Results of ECG-aBcDe on Various LLMs}
\label{tab:vllms}
\centering
\scalebox{1.0}{
\begin{tabular}{@{}lccccc@{}}
\toprule
\textbf{Model} & \textbf{Tr. Data} & \textbf{Inf. Data} & \textbf{BLEU-4} & \textbf{Rouge-L} & \textbf{Meteor} \\
\midrule
Gemma3 &  &  & \textbf{44.14} & \textbf{52.61} & \textbf{30.33} \\
Llama3.2 & PTB-XL & PTB-XL & \underline{42.58} & \underline{50.55} & \underline{29.32} \\
Qwen2.5 &  &  & 34.14 & 47.93 & 24.63 \\
\bottomrule
\end{tabular}
}
\end{table}

\section{Interpretability of Generated Outputs}
This section builds upon prior work \cite{ECG_Byte,jieshia} to further investigate attention-based visualization for enhancing the interpretability of generated outputs. Existing methods typically attend to sampling points, limiting their ability to interpret temporal information. To address this, we propose a "point–segment with attention" mechanism that attends not only to key points but also to the segments between them, thereby integrating time-scale information into interpretability. Unlike methods that visualize attention directly on original ECG \cite{jieshia}, our method visualizes attention weights on reconstructed ECG. Considering that ECG-aBcDe encoding is inherently lossy, the reconstructed ECG must maintain high morphological fidelity to the original. Otherwise, the attention computed on the reconstruction cannot accurately reflect the model’s true focus, reducing the reliability of the interpretation.

\begin{figure}[htbp]
    \centering
    \includegraphics[width=0.88\columnwidth]{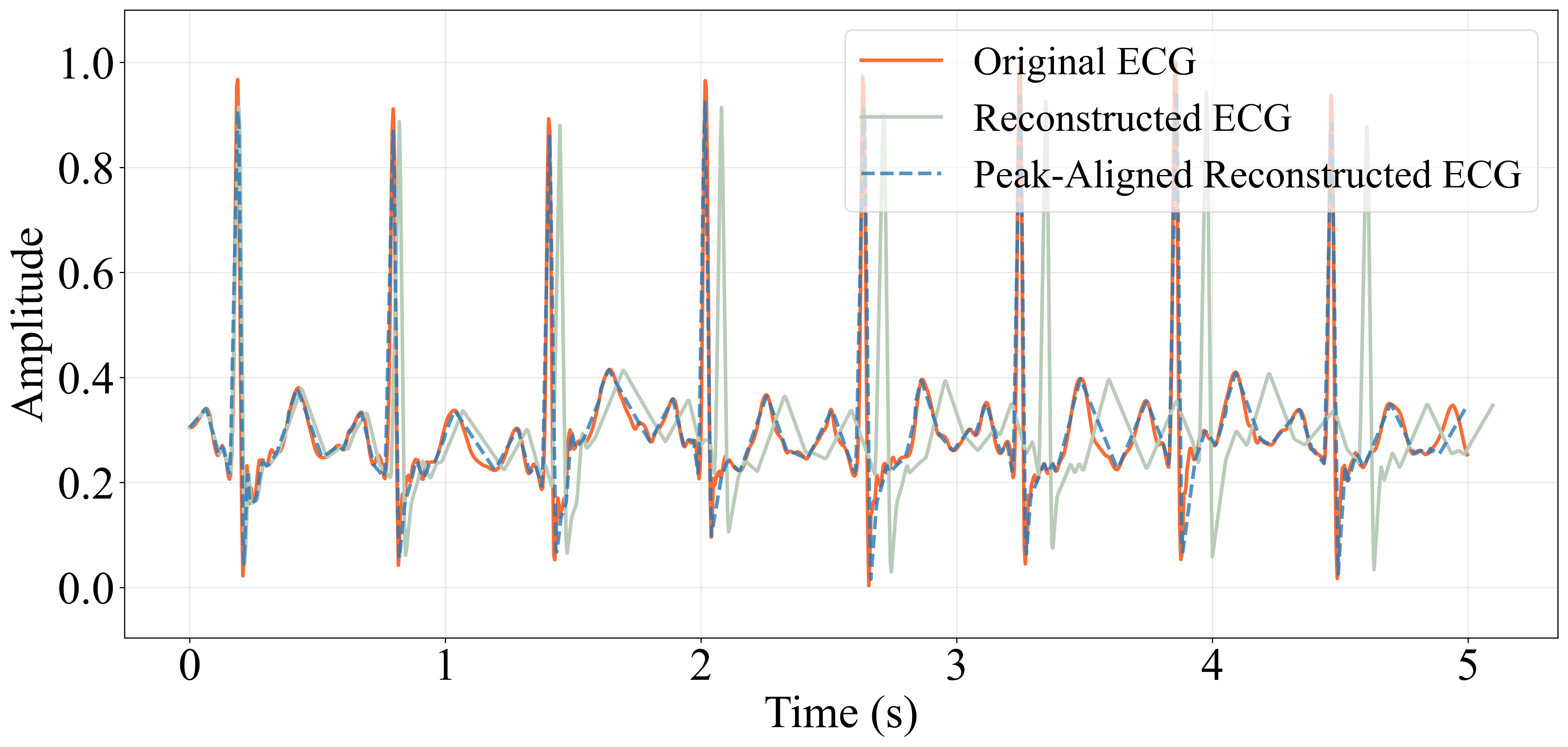}
    \caption{Visualization of original ECG, reconstructed ECG, and R-peak aligned reconstructed ECG.}
    \label{fig:leadi}
\end{figure}

To address this concern, we first evaluates the visual similarity between original and reconstructed ECGs. As illustrated in Figure \ref{fig:leadi}, accumulated errors during the encoding process can lead to time-axis stretching and rightward waveform shifts in the reconstructed signals. To mitigate these distortions, a peak-alignment-based resampling strategy is introduced, which effectively preserves both morphological features and temporal fidelity, thus laying a solid foundation for subsequent visualization-based interpretation.

\begin{figure}[htbp]
    \centering
    \includegraphics[width=0.88\columnwidth]{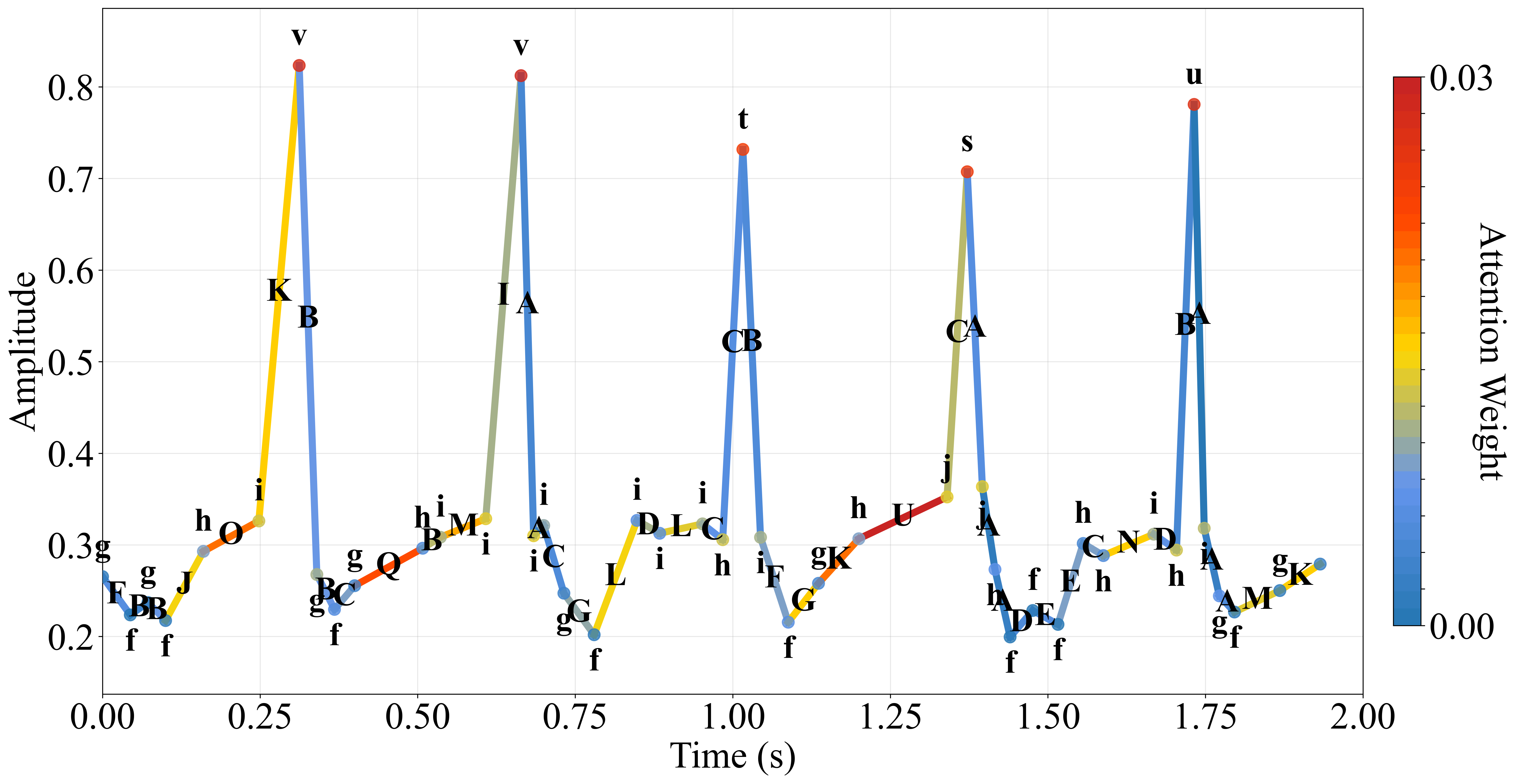}
    \caption{Visualization of Point–Segment with Attention.}
    \label{fig:letter}
\end{figure}

Based on the preceding analysis, this study visualizes the ECG signals using attention weights derived during the inference process of the large language model (LLM). Figure 4 presents the attention distribution when the model is queried with the question: “Does this ECG display any features of supraventricular tachycardia?” In this setting, the ECG string is first tokenized into multiple units such as 'g', 'F', 'Ah', and 'h'. Among these, lowercase letters correspond to key points, while uppercase letters denote the segments between adjacent key points. Attention intensities are encoded using color gradients: point-wise attention is reflected by the color of the respective markers, and segment-wise attention is visualized through the color of the connecting lines. This enables a comprehensive point-segment joint attention visualization, effectively capturing both discrete physiological events and the continuous temporal context within the ECG sequence. It is noteworthy that in Figure \ref{fig:letter}, temporal segments denoted by characters such as 'K', 'O', and 'Q', along with R-peaks represented by 's', 't', 'u', and 'v', received substantially higher attention weights than non-R-wave keypoints such as 'f' and 'h'. This observation suggests that the LLM identifies R-peaks and further analyzes RR intervals to determine the presence of supraventricular tachycardia (SVT). Such a mechanism closely aligns with clinical diagnostic reasoning, where SVT is commonly identified based on RR interval analysis, thereby enhancing the interpretability of the model’s generation process.

\section{Limitation of Large Language Models}
To evaluate the limitations of LLMs in counting, we constructed a synthetic fine-tuning dataset designed to isolate counting behavior. To minimize tokenizer interference, only the characters 'Z' and 'Q' were used. Empirical analysis confirmed that the tokenizer yields only three tokens including 'Z', 'Q', and 'ZZ'. The training set is constructed by generating $K \in {500,\ 1000,\ 1500,\ 2000}$ samples for each integer output value within the interval $(50,\ 150]$, resulting in a total of $100 \times K$ training samples. The test set follows the same sampling strategy, with 10 samples generated for each output value, yielding a total of 1000 samples for evaluating the model’s generalization performance.

To evaluate the limitations of LLMs in counting, we constructed a fine-tuning dataset specifically designed to isolate counting behavior. To eliminate tokenizer interference, only the characters 'Z' and 'Q' were used. Empirical analysis confirmed that the tokenizer produces only three token types: 'Z', 'Q', and 'ZZ', ensuring structural simplicity.
Each sample comprises an instruction, input, and output. The instruction is fixed as: "What is the minimum number of 'Z' found in any substring that starts and ends with 'Q'?" The input is a string of 500 'Z' characters, in which three positions $q_1 < q_2 < q_3$ are randomly chosen and replaced with 'Q'. The output is calculated as $\min(q_2 - q_1 - 1,\ q_3 - q_2 - 1)$, corresponding to the minimum number of 'Z's between adjacent 'Q's.
For each integer output value within the interval $(50, 150]$, $K \in \{500, 1000, 1500, 2000\}$ samples are generated, resulting in $100 \times K$ training samples. The test set is constructed independently using the same strategy, with 10 samples per output value, totaling 1000 examples for evaluating generalization.

\begin{figure}[htbp]
    \centering
    \includegraphics[width=0.88\linewidth]{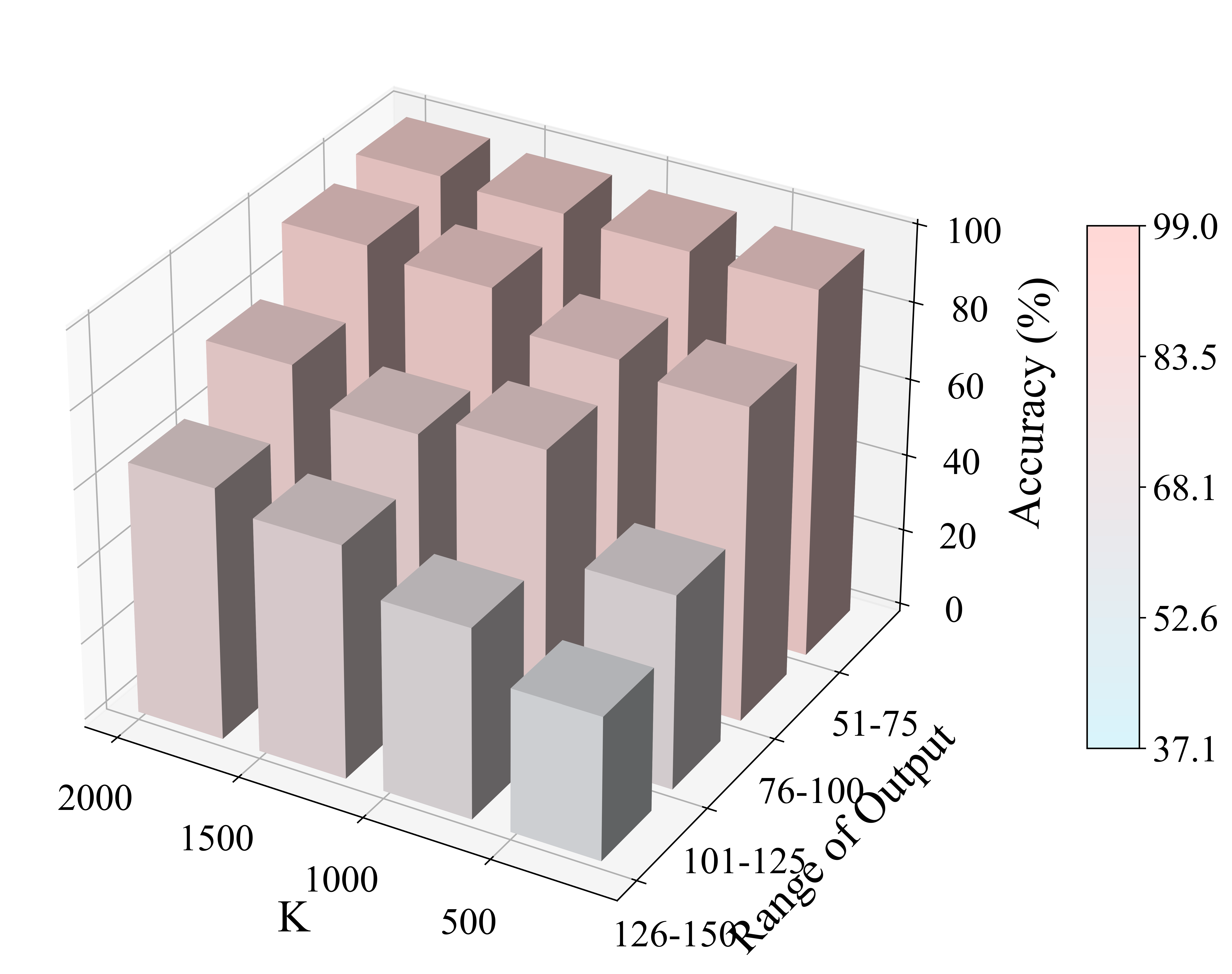} 
    \caption{Counting accuracy.}
    \label{fig:TransLimit}
\end{figure}

As shown in Figure \ref{fig:TransLimit}, LLaMA3.2-1B-Instruct exhibits a clear drop in counting accuracy as the distance between 'Q' tokens increases, revealing its difficulty in learning time-scale information. While increasing the training set size improves accuracy, the gains diminish rapidly: even with 200k samples, accuracy in the $(125,\ 150]$ range remains at just 65.15\%. This suggests a structural limitation in LLMs’ ability to model time-scale information, which cannot be resolved by data scaling alone, underscoring the need for explicit temporal encoding.

\section{Conclusion}
This study proposes the ECG-aBcDe encoding method, which transforms ECG signals into a natural language-like representation called "ECG language." This method enables large language models to efficiently learn and comprehend ECGs in a manner analogous to natural language understanding. ECG-aBcDe achieves a complete decoupling between ECG encoding and the LLM, overcoming limitations of existing methods requiring model-specific adaptations. Moreover, this study systematically identifies key bottlenecks current LLMs face when processing ECG signals. Experimental results demonstrate that LLMs exhibit deficiencies in capturing the time-scale information intrinsic to ECGs. To mitigate this issue, an explicit expression mechanism based on alternating sequences of key points and their inter-point intervals is introduced, partially addressing the model’s shortcomings in temporal reasoning. Additionally, the study enhances interpretability of outputs by introducing a novel “point–segment with attention” mechanism.

Despite promising results, several limitations remain. First, experiments were conducted only on LLMs with parameter sizes below 1B, and the performance of larger-scale models under the proposed encoding scheme remains unevaluated, leaving their potential unexplored. Second, although the proposed method provides effective interpretability for time-scale–related conditions such as tachycardia, it still falls short in explaining abnormal ECG patterns involving waveform ectopy or signal dropout. Future work will focus on integrating domain-specific medical knowledge to enhance the model’s capability in recognizing and understanding clinically critical features.

\bibliographystyle{unsrt}  
\bibliography{references}  






\end{document}